\documentclass[letterpaper]{article}
\usepackage{aaai}
\usepackage{times}
\usepackage{helvet}
\usepackage{courier}
\usepackage{amsmath}

\long\def\comment#1{}
\long\def\optional#1{\empty}

\def\ar{\leftarrow}
\def\beq{\begin{equation}}
\def\eeq#1{\label{#1}\end{equation}}
\def\ba{\begin{array}}
\def\ea{\end{array}}
\def\i#1{\hbox{\it #1\/}}

\def\no{\i{not}}

\def\sat{\models}

\def\ar{\leftarrow}
\def\rar{\rightarrow}

\def\no{\i{not}}
\def\sat{\models}

\def\i#1{\hbox{\itshape #1\/}}

\newtheorem{prop}{Proposition}
\newtheorem{thm}{Theorem}
\newtheorem{cor}{Corollary}
\newtheorem{lemma}{Lemma}

\newtheorem{definition}{Definition}

\def\proof{\noindent{\bf Proof}.\hspace{3mm}}
\def\qed{\quad \vrule height7.5pt width4.17pt depth0pt \medskip}

\comment{
\newenvironment{enumerate-p}{
\begin{enumerate}
  \setlength{\itemsep}{1pt}
  \setlength{\parskip}{0pt}
  \setlength{\parsep}{0pt}
}{\end{enumerate}}

\newenvironment{itemize-p}{
\begin{itemize}
  \setlength{\itemsep}{0.3em}
  \setlength{\parskip}{0pt}
  \setlength{\parsep}{0pt}
}{\end{itemize}}
}

\long\def\optional#1{\empty}

\title{\bf Elementary Sets for Logic Programs}

\author{
Martin Gebser \\
Institut f\"ur Informatik \\
Universit\"at Potsdam, Germany \\ 
\And 
Joohyung Lee \\ 
Computer Science and Engineering \\
Arizona State University, USA \\ 
\And
Yuliya Lierler \\
Department of Computer Sciences \\
University of Texas at Austin, USA
}

\begin{document}

\maketitle

\begin{abstract}
By introducing the concepts of a loop and a loop formula, 
Lin and Zhao showed that the answer sets of a nondisjunctive 
logic program are exactly the models of its Clark's completion 
that satisfy the loop formulas of all loops.  Recently, 
Gebser and Schaub showed that the Lin-Zhao theorem remains 
correct even if we restrict loop formulas to a special class 
of loops called ``elementary loops.'' In this paper, 
we simplify and generalize the notion of an elementary
loop, and clarify its role. We propose the notion of 
an elementary set, which is almost equivalent to the 
notion of an elementary loop for nondisjunctive programs, 
but is simpler, and, unlike elementary loops, can be 
extended to disjunctive programs without producing 
unintuitive results. We show that the maximal unfounded 
elementary sets for the ``relevant'' part of a program are 
exactly the minimal sets among the nonempty unfounded 
sets. We also present a graph-theoretic 
characterization of elementary sets for nondisjunctive 
programs, which is simpler than the one proposed in~\cite{gebsch05a}.
Unlike the case of nondisjunctive programs, we show that
the problem of deciding an elementary set is 
{\sf coNP}-complete for disjunctive programs. \\[-1em]
\end{abstract}

\vspace{-1em}
\section{Introduction} \label{sec:intro}

By introducing the concepts of a loop and a loop formula, 
Lin and Zhao~\shortcite{lin04} showed that the answer sets
(a.k.a. stable models) of a nondisjunctive logic program are 
exactly the models of its Clark's completion~\cite{cla78} that satisfy 
the loop formulas $\i{LF}(L)$ of all loops $L$ for the program. 
This important result has shed new light on the relationship between 
answer sets and completion, and allowed us to compute
answer sets using SAT solvers, which led to the design of 
answer set solvers
{\sc assat}\footnote{{\tt http://assat.cs.ust.hk/}} \cite{lin04}
and 
{\sc cmodels}\footnote{{\tt http://www.cs.utexas.edu/users/tag/cmodels/}}
\cite{giu04a}.

The concepts of a loop and a loop formula were further clarified 
in~\cite{lee05}. 
By slightly modifying the definition of a loop, 
Lee observed that adding loop 
formulas can be viewed as a generalization of completion, which allows 
us to characterize the stability of a model in terms of loop formulas: 
A model is stable iff it satisfies the loop formulas of all loops.
He also observed that the mapping $\i{LF}$, which turns loops 
into loop formulas, can be applied to arbitrary 
sets of atoms, not only to loops: Adding $\i{LF}(Y)$ for a non-loop $Y$
does not affect the models of the theory because 
$\i{LF}(Y)$ is always entailed by $\i{LF}(L)$ for some loop $L$. 
Though this reformulation of the Lin-Zhao theorem, in which 
$\i{LF}$ is not restricted to loops, is less economical, 
it is interesting to note that it is essentially a theorem 
on assumption sets~\cite{sac90}, or unfounded 
sets~\cite{van91,lerusc97a}, which has been known for many years.
In this sense, the most original contribution 
of~\cite{lin04} was not the mapping that turns loops into 
loop formulas, but the definition of a loop, which yields a 
relatively small class of sets of atoms for the mapping~$\i{LF}$.

However, for nondisjunctive programs, even the definition of a loop 
turned out still ``too generous.''
Gebser and Schaub~\shortcite{gebsch05a} showed that 
restricting the mapping  even more to a special class of loops called 
``elementary loops,'' yields a valid modification of 
the Lin-Zhao theorem (or the Sacc\'a-Zaniolo theorem). 
That is, some loops are identified as redundant, just as all
non-loops are redundant. 
They noted that the notion of a positive dependency graph, 
which is used for defining a loop, is not expressive enough 
to 
distinguish between elementary and non-elementary loops, 
and instead proposed 
another graph-theoretic characterization,
based on the notion of a so-called 
``body-head dependency graph.''

Our work is motivated by the desire to understand the role of an 
elementary loop further and to extend the results to disjunctive programs.
For nondisjunctive programs, we propose a simpler notion corresponding 
to an elementary loop, which we call an ``elementary set,'' and provide 
a further enhancement of the Lin-Zhao theorem based on it.  
Unlike elementary loops, elementary sets can be extended to 
disjunctive programs without producing unintuitive results. 
We show that a special class of unfounded elementary sets coincides with
the minimal sets among nonempty unfounded sets.
Instead of relying on the notion of a body-head dependency 
graph, we present a simpler graph-theoretic characterization of 
elementary sets, based on a subgraph of the positive 
dependency graph.

\vspace{-0.7em}
\section{Nondisjunctive Programs} \label{sec:elm-nondis}

\vspace{-0.3em}
\subsection{Review of Loop Formulas: Nondisjunctive Case}
\vspace{-0.3em}

A {\sl nondisjunctive rule} is an expression of the form
\pagebreak
\beq
  a_1 \ar a_2,\dots,a_m,\no\ a_{m+1},\dots,\no\ a_n
\eeq{nd}
where $n\geq m\geq 1$ and $a_1,\dots,a_n$ are propositional atoms.
A {\sl nondisjunctive program} is a finite set of nondisjunctive rules.  

We will identify a nondisjunctive rule~(\ref{nd}) with the 
propositional formula
\[ 
(a_2\land\cdots\land a_m\land\neg a_{m+1}\land\cdots\land\neg a_n)
                     \rar a_1, 
\] 
and will often write~(\ref{nd}) as
\beq
  a_1\ar B, F
\eeq{abf-nd}
where $B$ is $a_2,\dots,a_m$ and $F$ is $\no\ a_{m+1},\dots,\no\ a_n$. 
We will sometimes identify $B$ with its corresponding set.

For the definition of a stable model of a nondisjunctive program, 
we refer the reader to~\cite[Section~2.1]{lee05}. 

Let $\Pi$ be a nondisjunctive program. 
The {\sl (positive) dependency graph} of~$\Pi$ is the directed graph 
such that its vertices are the atoms occurring in~$\Pi$, and
its edges go from~$a_1$ to~$a_2,\dots, a_m$ for all rules~(\ref{nd}) 
of~$\Pi$.
A nonempty set~$L$ of atoms is called a {\sl loop} of~$\Pi$ if, for every 
pair~$p$, $q$ of atoms in~$L$, there exists a path (possibly of length~$0$)
from~$p$ to $q$ in the dependency graph of~$\Pi$ such that all vertices 
in this path belong to~$L$.  
In other words, $L$ is a loop of~$\Pi$ iff 
the subgraph of the dependency graph of~$\Pi$ induced by~$L$ is strongly
connected.  
Clearly, any set consisting of a single atom is a loop.
For example, the following program $\Pi_1$ 
\[ 
\ba {l}
p \ar \no\ s  \hspace{1.5em}
p \ar r  \hspace{1.5em} 
q \ar r  \hspace{1.5em}
r \ar p,q  \ 
\ea
\] 
has seven loops: $\{p\}$, $\{q\}$, $\{r\}$, $\{s\}$,
$\{p,r\}$, $\{q,r\}$, $\{p, q, r\}$. 

For any set $Y$ of atoms, the {\sl external support formula} of~$Y$
for~$\Pi$, denoted by $\i{ES}_\Pi(Y)$, is the disjunction of conjunctions  
$  B\land F  $ 
for all rules~(\ref{abf-nd}) of~$\Pi$ such that $a_1\in Y$~and  
$B\cap Y = \emptyset$.
The first condition expresses that the atom ``supported'' by~(\ref{abf-nd}) 
is an element of~$Y$. The second condition ensures that this support 
is ``external'': The atoms in~$B$ that it relies on do not belong
to~$Y$. Thus $Y$ is called {\sl externally supported} by $\Pi$ w.r.t.
a set~$X$ of atoms if $X$ satisfies~$\i{ES}_\Pi(Y)$.\footnote{We identify 
an interpretation with the set of atoms that are true in it. }

For any set $Y$ of atoms, by $\i{LF}_\Pi(Y)$ we denote the following 
formula: 
\beq
   \mbox{$\bigwedge$}_{a\in Y} a \rar \i{ES}_\Pi(Y) \ .
\eeq{lf}
Formula~(\ref{lf}) is called the {\sl (conjunctive) loop formula} of~$Y$
for~$\Pi$.\footnote{If the conjunction in the antecedent is replaced with 
the disjunction, the formula is called 
{\sl disjunctive loop formula} \cite{lin04}. Our results stated 
in terms of conjunctive loop formulas can be stated in terms 
of disjunctive loop formulas as well.}
Note that we still call (\ref{lf}) a loop formula even when $Y$ is not a loop.
The following reformulation of the Lin-Zhao theorem, which characterizes 
the stability of a model in terms of loop formulas, is a part of 
the main theorem from~\cite{lee05} for the nondisjunctive case.

\begin{thm}~\cite{lee05} \label{thm:lf}
Let~$\Pi$ be a nondisjunctive program, and $X$ a set of atoms
occurring in~$\Pi$. If $X$ satisfies~$\Pi$, 
then the following conditions are equivalent: 
\begin{description}
\item[\textnormal{\textit{(a)}}]  
   $X$ is stable;
\item[\textnormal{\textit{(b)}}]  
   $X$ satisfies~$\i{LF}_\Pi(Y)$ for all nonempty sets $Y$ 
   of atoms that occur in~$\Pi$;            
\item[\textnormal{\textit{(c)}}]  
   $X$ satisfies~$\i{LF}_\Pi(Y)$ for all loops $Y$ of~$\Pi$.
\end{description}
\end{thm}

According to the equivalence between conditions~(a) and~(b) in
Theorem~\ref{thm:lf}, a model of~$\Pi_1$ is stable iff it satisfies 
the loop formulas of all fifteen nonempty sets of atoms occurring in $\Pi_1$. 
On the other hand, condition~(c) tells us that it is sufficient 
to restrict attention to the following seven loop formulas:
\beq
\ba {lllll}
   p \rar \neg s \lor r &\hspace{0.2em}&
       r \rar  p \land q  &\hspace{0.5em}&
            p \land r \rar \neg s  \\
   q \rar  r  & &
       s \rar \bot  & &
            q \land r \rar \bot \\
         &  & 
         &  &
       p \land q \land r \rar \neg s \ . \\
\ea
\eeq{ex-lf}
Program~$\Pi_1$ has six models: $\{p\}$, $\{s\}$, $\{p,s\}$, $\{q,s\}$,   
$\{p,q,r\}$, and $\{p,q,r,s\}$. Among them, $\{p\}$ is the only 
stable model, which is also the only model that satisfies all loop 
formulas~(\ref{ex-lf}). In the next section, we will see 
that in fact the last loop formula can be disregarded as well, 
if we take elementary sets into account.

As noted in~\cite{lee05}, the equivalence between conditions~(a) and (c)
is a reformulation of the Lin-Zhao theorem;  the equivalence between 
conditions~(a) and (b) is a reformulation of Corollary~2 of \cite{sac90}, 
and Theorem~4.6 of \cite{lerusc97a} (for the nondisjunctive case), 
which characterizes the stability of a model in terms of 
{\sl unfounded sets}. 
For sets $X$, $Y$ of atoms, we say that $Y$ is {\sl unfounded} by~$\Pi$ 
w.r.t.~$X$ if $Y$ is not externally supported by~$\Pi$ w.r.t.~$X$.
Condition~(b) can be stated in terms of unfounded sets as follows: 
{\it
\begin{description}
\item[\textnormal{\textit{(b$'$)}}] 
     $X$ contains no nonempty unfounded subsets for~$\Pi$ w.r.t.~$X$.
\end{description}
}

\vspace{-0.7em}
\subsection{Elementary Sets for Nondisjunctive Programs}
\vspace{-0.2em}

As mentioned in the introduction, \cite{gebsch05a} showed that $\i{LF}$ in
Theorem~\ref{thm:lf} can be further restricted to ``elementary loops.''
In this section, we present a simpler reformulation of their results. 
We will compare our reformulation with the original definition 
from~\cite{gebsch05a} later in this paper. 

Let $\Pi$ be a nondisjunctive program.
The following proposition tells us that a loop can be defined even 
without referring to a dependency graph.
\begin{prop}\label{prop:loop-alternative}
For any nondisjunctive program~$\Pi$ and any nonempty set 
$Y$ of atoms occurring in~$\Pi$, $Y$ is a loop of~$\Pi$ iff, 
for every nonempty proper subset $Z$ of $Y$, there is 
a rule~(\ref{abf-nd}) in~$\Pi$ such that 
$a_1\in Z$ and $B\cap (Y\setminus Z)\ne\emptyset$.
\end{prop}

For any set $Y$ of atoms and any subset $Z$ of $Y$, we say that 
$Z$ is {\sl outbound} in $Y$ for $\Pi$ 
if there is a rule~(\ref{abf-nd}) in~$\Pi$ such that 
$a_1\in Z$,\;  $B\cap (Y\setminus Z)\ne\emptyset$, and
$B\cap Z=\emptyset$.

For any nonempty set $Y$ of atoms that occur in $\Pi$, we say that 
$Y$ is {\sl elementary} for~$\Pi$ if all nonempty proper subsets 
of~$Y$ are outbound in $Y$ for $\Pi$. 

As with loops, it is clear from the definition that every set
consisting of a single atom occurring in $\Pi$ is elementary for~$\Pi$.
It is also clear that every elementary set for $\Pi$ is a loop of
$\Pi$, but a loop is not necessarily an elementary set: 
The conditions for being an elementary set are stronger than 
the conditions for being a loop as given in 
Proposition~\ref{prop:loop-alternative}.
For instance, one can check that for $\Pi_1$, $\{p,q,r\}$  is not 
elementary since $\{p,r\}$ (or $\{q,r\}$) is not outbound in 
$\{p,q,r\}$. 
All the other loops of $\Pi_1$ are elementary. Note that an elementary set 
may be a proper subset of another elementary set (both $\{p\}$ and $\{p,r\}$
are elementary sets for $\Pi_1$). 

\comment{
The following program replaces the last rule of~$\Pi_1$ by two rules: \\[-0.7em]
\beq 
\ba {lllll}
  p \ar \no\ s  &\hspace{2em}&
      q \ar r   &\hspace{2em}&
           r \ar q . \\
  p \ar r  &\hspace{2em}&
      r \ar p
\ea
\eeq{ex2}
Program~(\ref{ex2}) has the same dependency graph as program~$\Pi_1$ and 
hence has the same set of loops. However its elementary sets 
are different: All its loops are elementary. \\[-0.7em]
}

\comment{
Consider another program:
\beq
\ba {lllll}
  p \ar r  &\hspace{2em}&
      q \ar q, r   &\hspace{2em}&
           r \ar p, q \\
  p \ar \no\ s  &\hspace{2em}&
      q \ar s      &\hspace{2em}&
           s \ar q, s \ .
\ea
\eeq{ex4}
Program~(\ref{ex4}) has $10$ loops: $\{p\}$, $\{q\}$, $\{r\}$, $\{s\}$, 
$\{p,r\}$, $\{q,r\}$, $\{q,s\}$, $\{p,q,r\}$, $\{q,r,s\}$, $\{p,q,r,s\}$. 
Among them, only the first five sets are elementary.
}

 From the definition of an elementary set above, we get an alternative, 
equivalent definition by requiring that only the loops contained in~$Y$ 
be outbound, instead of requiring that all nonempty proper subsets
of~$Y$ be outbound.
\begin{prop} \label{prop:elm-def}
For any nondisjunctive program~$\Pi$ and any nonempty set $Y$ 
of atoms that occur in~$\Pi$, $Y$ is an elementary set for~$\Pi$ 
iff all loops $Z$ of $\Pi$ such that $Z\subset Y$ are outbound 
in $Y$ for $\Pi$.\footnote{Note that Proposition~\ref{prop:elm-def} 
remains correct even after replacing ``all loops'' in its statement 
with ``all elementary sets.''}
\end{prop}

Note that a subset of an elementary set, even if that subset is a loop, 
is not necessarily elementary. For instance, for program
\\[-1.0em]
\begin{equation*}
\ba {lllll}
~~~~~~~ p \ar p, q &\hspace{2em}&
     p \ar r  &\hspace{2em}&    
          r \ar p \\
~~~~~~~ q \ar p, q   &\hspace{2em}&
     q \ar r &\hspace{2em}&
          r \ar q, 
\ea
\end{equation*}
set $\{p,q,r\}$ is elementary, but $\{p,q\}$ is not.

The following proposition describes a relationship between loop formulas
of elementary sets and those of arbitrary sets.
\begin{prop} \label{prop:elm-entail}
Let $\Pi$ be a nondisjunctive program, $X$ a set of atoms, and 
$Y$ a nonempty set of atoms that occur in $\Pi$. 
If $X$ satisfies $\i{LF}_\Pi(Z)$ for all elementary sets $Z$ of $\Pi$
such that $Z\subseteq Y$, then $X$ satisfies $\i{LF}_\Pi(Y)$.
\end{prop}

Proposition~\ref{prop:elm-entail} suggests that condition (c) of 
Theorem~\ref{thm:lf} can be further enhanced by taking only loop 
formulas of elementary sets into account. This yields the following
theorem, which is a reformulation of Theorem~3
from~\cite{gebsch05a} in terms of elementary sets.

\medskip
\noindent{\bf Theorem~\ref{thm:lf}(d)}\ \
{\it 
The following condition is equivalent to conditions~(a)--(c) of 
Theorem~\ref{thm:lf}.
\begin{description}
\item[\textnormal{\textit{(d)}}]
      $X$ satisfies~$\i{LF}_\Pi(Y)$ for all elementary sets~$Y$ of~$\Pi$. 
\end{description}
}

According to Theorem~\ref{thm:lf}(d), a model of~$\Pi_1$ is stable iff 
it satisfies the first six formulas in~(\ref{ex-lf}); the loop formula 
of non-elementary set $\{p,q,r\}$ (the last one in~(\ref{ex-lf})) 
can be disregarded.

\vspace{-0.2em}
\subsection{
Elementarily Unfounded Sets for Nondisjunctive Programs} 
\vspace{-0.2em}

If we modify condition~(c) of Theorem~\ref{thm:lf} by replacing
``loops'' in its statement with ``maximal loops,'' the condition
becomes weaker, and the modified statement of Theorem~\ref{thm:lf} 
does not hold. For instance, program $\Pi_1$ has only two maximal loops, 
$\{p,q,r\}$ and $\{s\}$, and their loop formulas are satisfied
by the non-stable model $\{p,q,r\}$. In fact, maximal loop $\{p,q,r\}$
is not even an elementary set for $\Pi_1$.

This is also the case with maximal elementary sets:
Theorem~\ref{thm:lf}(d) does not hold if ``elementary sets'' in its 
statement is replaced with ``maximal elementary sets''
as the following program shows: 
\beq
\ba {lll}
 p \ar q, \no\ p \hspace{2.5em}
 q \ar p, \no\ p \hspace{2.5em}
 p \ .
\ea
\eeq{ex3}
Program~(\ref{ex3}) has two models, $\{p\}$ and $\{p,q\}$, but the latter 
is not stable. Yet, both models satisfy the 
loop formula of the only maximal elementary set $\{p,q\}$ 
for $(\ref{ex3})$ ($p\land q\rar \top$).

However, in the following we show that if we consider 
the ``relevant'' part of the program w.r.t.\ a given interpretation, 
it is sufficient to restrict attention to maximal elementary sets. 

Given a nondisjunctive program $\Pi$ and a set $X$ of atoms,
by $\Pi_X$ we denote the set of rules~(\ref{abf-nd}) of~$\Pi$ 
such that \hbox{$X\sat B,F$}. 
The following proposition tells 
us that all nonempty proper subset of an elementary set
for~$\Pi_X$ are externally supported w.r.t.~$X$.

\begin{prop} \label{prop:max-elm}
For any nondisjunctive program~$\Pi$, any set~$X$ of atoms, and 
any elementary set~$Y$ for $\Pi_X$, $X$ satisfies $\i{ES}_\Pi(Z)$
for all nonempty proper subsets $Z$ of $Y$.
\end{prop}

 From Proposition~\ref{prop:max-elm}, it follows that every unfounded 
elementary set~$Y$ for~$\Pi_X$ w.r.t.~$X$ is maximal among 
the elementary sets for~$\Pi_X$.
One can show that if $Y$ is a nonempty unfounded set for~$\Pi$ w.r.t.~$X$
that does not contain a maximal elementary set for~$\Pi_X$, 
then~$Y$ consists of atoms that do not occur in $\Pi_X$.
 From this, we obtain the following result. 

\medskip
\noindent{\bf Theorem~\ref{thm:lf}(e)}\ \
{\it 
The following condition is equivalent to conditions~(a)--(c) of 
Theorem~\ref{thm:lf}.
\begin{description}
\item[\textnormal{\textit{(e)}}]  
   $X$ satisfies $\i{LF}_\Pi(Y)$ for every set $Y$ of atoms 
   such that $Y$ is a maximal elementary set for $\Pi_X$, or
   a singleton whose atom occurs in~$\Pi$.
\end{description}
}

\comment{
According to Theorem~\ref{thm:lf}(e), a model $\{p\}$ of $\Pi_1$ 
is stable because atom $p$ occurs in $(\Pi_1)_{\{p\}} = \{p \ar \no\ r\}$, 
and satisfies the external support formula $\neg r$.
On the other hand, a model $\{p,q,r\}$ of $\Pi_1$ is not stable 
because it does not satisfy the external support formulas for the 
maximal elementary sets for $(\Pi_1)_{\{p,q,r\}}$, which are 
$\{p,r\}$ and $\{q,r\}$.
}

\comment{
Note that the analogy does not apply to loops: If we replace
``maximal elementary sets'' in the statement of Theorem~\ref{thm:lf}(e)
with ``maximal loops,'' then the modified statement does not hold.
The non-stable model $\{p,q,r\}$ still satisfies the loop formula of 
the maximal loop of $(\Pi_1)_{\{p,q,r\}}$ (the last one in~(\ref{ex-lf})).
}

We say that a set $Y$ of atoms occurring in $\Pi$ is 
{\sl elementarily unfounded} by $\Pi$ w.r.t.~$X$ if 
$Y$ is an elementary set for~$\Pi_X$ that is unfounded 
by~$\Pi$ w.r.t.~$X$, or $Y$ is a singleton that is unfounded by 
$\Pi$ w.r.t.~$X$.\footnote{Elementarily
unfounded sets are closely related to ``active elementary loops'' 
in~\cite{gebsch05a}.}
 From Proposition~\ref{prop:max-elm}, 
every non-singleton
elementarily unfounded set for~$\Pi$ w.r.t.~$X$ is a maximal
elementary set for~$\Pi_X$.

It is clear from the definition that every elementarily unfounded set 
for~$\Pi$ w.r.t.~$X$ is an elementary set for~$\Pi$ and that it is also 
an unfounded set for~$\Pi$ w.r.t.~$X$. However, a set that is both
elementary for~$\Pi$ and unfounded by~$\Pi$ w.r.t.~$X$ is not necessarily
an elementarily unfounded set for~$\Pi$ w.r.t.~$X$.
For example, consider the following program: 
\beq
 p \ar q, \no\ r \hspace{2.5em}
 q \ar p, \no\ r  \ .
\eeq{ex-el-uf}
Set $\{p,q\}$ is both elementary for (\ref{ex-el-uf}), 
and unfounded by (\ref{ex-el-uf}) w.r.t.~$\{p,q,r\}$, but it is not an
elementarily unfounded set w.r.t.~$\{p,q,r\}$.

The following corollary, which follows from Proposition~\ref{prop:max-elm}, 
tells us that all nonempty proper subsets of an elementarily 
unfounded set are externally supported. It is essentially 
a reformulation of~Theorem~5 from~\cite{gebsch05a}.
\begin{cor} \label{cor:elm-uf}
Let $\Pi$ be a nondisjunctive program, $X$ a set of atoms, and $Y$ 
an elementarily unfounded set for~$\Pi$ w.r.t.~$X$.  Then
$X$ does not satisfy $\i{ES}_\Pi(Y)$, but
satisfies $\i{ES}_\Pi(Z)$ for all nonempty proper 
subsets~$Z$ of~$Y$.
\end{cor}

Corollary~\ref{cor:elm-uf} tells us that elementarily unfounded sets 
form an ``anti-chain'': One of them cannot be a proper subset of 
another.\footnote{Recall that the anti-chain property 
does not hold for elementary sets for $\Pi$: An elementary set may 
contain another elementary set as its proper subset.} 
In combination with Proposition~\ref{prop:max-elm}, this tells us that 
elementarily unfounded sets are minimal among nonempty 
unfounded sets. Interestingly, the converse also holds. 

\begin{prop} \label{prop:min-uf}
For any nondisjunctive program $\Pi$ and any sets $X$, $Y$ of atoms, 
$Y$ is an elementarily unfounded set for~$\Pi$ w.r.t.~$X$ iff 
$Y$ is minimal among the nonempty sets of atoms occurring in~$\Pi$
that are unfounded by~$\Pi$ w.r.t.~$X$.
\end{prop}

Theorem~\ref{thm:lf}(e) can be stated in terms of elementarily unfounded 
sets, thereby restricting attention to minimal unfounded sets.\\[-1em]

{\it
\begin{description}
\item[\textnormal{\textit{(e$'$)}}]  
  $X$ contains no elementarily unfounded subsets for~$\Pi$
  w.r.t.~$X$.\\[-1em]
\end{description}
}

The notion of an elementarily unfounded set may help improve
computation performed by SAT-based answer set solvers.
Since there are exponentially many loops in the worst case, 
SAT-based answer set solvers do not add all loop formulas at once.
Instead, they check whether a model returned by a SAT solver is an answer 
set.
If not, a loop formula that is not satisfied by the current
model is added, and the SAT solver is invoked again.\footnote{%
To be precise, {\sc cmodels} adds ``conflict clauses.''}
This process is repeated until an answer set is found,
or the search space is exhausted. 
In view of Theorem~\ref{thm:lf}(e$'$), when loop formulas need
to be added, it is sufficient to add loop formulas of elementarily 
unfounded sets only. 
This guarantees that loop formulas considered are only those of 
elementary sets. Since every elementary set is a loop, but
not vice versa, the process may involve fewer loop formulas 
overall than the case when arbitrary loops are considered.
In view of Proposition~\ref{prop:elm-entail} and 
Corollary~\ref{cor:elm-uf}, this would yield reasonably the most 
economical way to eliminate unfounded models.

\optional{
\proof
Follows from Lemma~\ref{lem:nob-entail} 
and Proposition~\ref{prop:max-elm}.
\qed
}

\vspace{-0.2em}
\subsection{Deciding Elementary Sets: Nondisjunctive Case}
\vspace{-0.2em}

The above definition of an elementary set  
involves all its nonempty proper subsets (or at least all loops that 
are its subsets). This seems to imply that deciding
whether a set is elementary is a computationally hard problem.
But in fact, \cite{gebsch05a} showed
that, for nondisjunctive programs, deciding an elementary loop
can be done efficiently. 
They noted that positive dependency graphs are not expressive 
enough to
distinguish between elementary and non-elementary loops, and instead 
introduced so-called ``body-head dependency graphs'' 
to identify elementary loops.
In this section, we simplify this result by 
still referring to positive dependency graphs.
We show that removing some ``unnecessary'' edges 
from the dependency graph is just enough to distinguish
elementary sets from non-elementary sets.

\begin{figure}
\begin{center}
 \begin{picture}(120,10)(0,3)
 \put(0,0){$p$}
 \put(90,0){$q$}
 \put(45.5,0){$r$}

 \qbezier(8,5)(25,11)(42,5) \put(43,4.6){\vector(3,-1){1}}

 \qbezier(88,5)(71,11)(54,5) \put(53,4.6){\vector(-3,-1){1}}
 \end{picture}
\caption{The elementary subgraph of $\{p,q,r\}$ for $\Pi_1$}
\label{fig:dep2}
\end{center}
\end{figure}
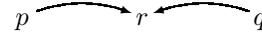

For any nondisjunctive program $\Pi$ and any set $Y$ of atoms,
\vspace{-1em}

\[
\ba l
\i{EC}_\Pi^0(Y) = \emptyset\ , \\
\i{EC}_\Pi^{i+1}(Y) = \i{EC}^i_\Pi(Y)\ \cup \{ (a_1,b) \mid
\text{there is a rule~(\ref{abf-nd}) in $\Pi$}  \\
  \hspace{2em}  \text{such that $b\in B$ and the graph
  $(Y,\i{EC}_\Pi^i(Y))$ has a} \\
  \hspace{2em} \text{strongly connected subgraph containing all atoms}  \\
  \hspace{2em} \text{in $B\cap Y$} \} \ , \\
\i{EC}_\Pi(Y) = \mbox{$\bigcup$}_{i\ge 0} \i{EC}_\Pi^i(Y) \ . 
\ea
\]
Note that this is a ``bottom-up'' construction. 
We call the graph $(Y,\i{EC}_\Pi(Y))$ the {\sl elementary subgraph}
of $Y$ for $\Pi$. It is clear that an elementary subgraph is a subgraph of 
a dependency graph and that it is not necessarily the same as the 
subgraph of the dependency graph induced by $Y$. 
Figure~\ref{fig:dep2} shows the elementary subgraph of~$\{p,q,r\}$ 
for~$\Pi_1$, which is not strongly connected. 

The following theorem is similar to \cite[Theorem~10]{gebsch05a}, but
instead of referring to the notion of a body-head dependency graph, 
it refers to an elementary subgraph as defined above.
\begin{thm} \label{thm:ec-tr}
For any nondisjunctive program $\Pi$ and any set~$Y$ of atoms
occurring in $\Pi$, $Y$ is an elementary set for~$\Pi$ iff 
the elementary subgraph of $Y$ for $\Pi$ is strongly connected.
\end{thm}

Clearly, constructing an elementary subgraph and checking whether it
is strongly connected can be done in polynomial time.
Therefore, the problem of deciding whether a given set of atoms is
elementary is tractable.

\section{Disjunctive Programs}
   \label{sec:elm-dis}
\subsection{Review of Loop Formulas: Disjunctive Case}

A {\sl disjunctive rule} is an expression of the form  \\[-0.7em]
\beq
\ba {rcl}
  a_1;\dots; a_k &\hspace{-1em}\ar\hspace{-1em}&  
                    a_{k+1},\dots, a_l, \no\ a_{l+1},\dots, \no\ a_m, \\
                 &     & \hspace{2em} \no\ \no\ a_{m+1},\dots, \no\ \no\ a_n
\ea
\eeq{dis}
where $n\ge m\ge l\ge k\ge 0$ and $a_1,\dots, a_n$ are propositional atoms. 
A {\sl disjunctive program} is a finite set of disjunctive rules. 

We will identify a disjunctive rule~(\ref{dis}) with the propositional 
formula 
\[
\ba l
  (a_{k+1}\land\cdots\land a_l\land\neg a_{l+1}\land\cdots\land\neg a_m
  \land \\ 
   \hspace{5em} \neg\neg a_{m+1}\land\cdots\land\neg\neg a_n) \rar (a_1\lor\cdots\lor a_k) \ , 
\ea
\comment{
\footnote{Although double negations can be eliminated in propositional 
logic without changing the models, it plays a different role 
under the stable model semantics~\cite[Section~2.1]{fer05e}.}
}
\]
and will often write~(\ref{dis}) as \\[-1em]
\beq
  A \ar B, F
\eeq{abf-dis}
where $A$ is $a_1,\dots,a_k$, $B$ is $a_{k+1},\dots,a_l$, and $F$ is 
$$
   \no\ a_{l+1},\dots,\no\ a_m,\no\ \no\ a_{m+1},\dots,\no\ \no\ a_n.
$$
We will sometimes identify $A$ and $B$ with their corresponding sets.

For the definition of a stable model of a disjunctive program, 
we refer the reader to~\cite[Section~2.2]{lee05}. 

The definition of a dependency graph 
is extended to a disjunctive program in a straightforward way: 
The vertices of the graph are the atoms occurring in the program, 
and its edges go from the elements of~$A$ to the elements of~$B$ 
for all rules (\ref{abf-dis}) of the program.
The definition of a loop in terms of the dependency graph remains 
the same as in the case of nondisjunctive programs.

Let $\Pi$ be a disjunctive program. For any set $Y$ of atoms, the
{\sl external support formula} of~$Y$ for~$\Pi$, denoted
by~$\i{ES}_\Pi(Y)$, is the disjunction of conjunctions 
\[ 
    B\land F\land \mbox{$\bigwedge$}_{a\in A\setminus Y} \neg a 
\]
for all rules~(\ref{abf-dis}) of~$\Pi$ such that
\hbox{$A\cap Y\ne\emptyset$} and \hbox{$B\cap Y = \emptyset$}.
When $\Pi$ is nondisjunctive, this definition reduces 
to the definition of $\i{ES}_\Pi$ for nondisjunctive programs 
given earlier.

The notion of $\i{LF}_\Pi$ and the term {\sl (conjunctive) loop 
formula} similarly apply to formulas~(\ref{lf}) when $\Pi$ is 
a disjunctive program.
As shown in \cite{lee05}, Theorem~\ref{thm:lf} remains correct 
after replacing ``nondisjunctive program'' in its statement 
with ``disjunctive program.''

\vspace{-0.2em}
\subsection{Elementary Sets for Disjunctive Programs}

In this section, we generalize the definition of an elementary set
to disjunctive programs. 

Note that a loop of a disjunctive program can be also defined 
without referring to a dependency graph:
Proposition~\ref{prop:loop-alternative} remains correct after replacing 
``nondisjunctive'' in its statement with ``disjunctive,'' 
``(\ref{abf-nd})'' with ``(\ref{abf-dis}),''
and ``$a_1\in Z$'' with ``$A\cap Z\ne\emptyset$.''

\comment{
\medskip
\noindent{\bf Proposition~\ref{prop:loop-alternative}$'$}\ \
{\it 
For any disjunctive program~$\Pi$ and for any nonempty set $Y$ of atoms 
that occur in~$\Pi$, $Y$ is a loop of~$\Pi$ iff, for every nonempty 
proper subset $Z$ of $Y$, there is a rule~(\ref{abf-dis}) in~$\Pi$ 
such that 
\begin{itemize}
\item  $A\cap Z\ne\emptyset$, and
\item  $B\cap (Y\setminus Z)\ne\emptyset$.
\end{itemize} 
}\medskip
}

Let $\Pi$ be a disjunctive program.  For any set $Y$ of atoms, 
we say that a subset $Z$ of $Y$ is {\sl outbound} in $Y$ for $\Pi$ 
if there is a rule~(\ref{abf-dis}) in~$\Pi$ such that 
$A\cap Z\ne\emptyset$, $B\cap (Y\setminus Z)\ne\emptyset$, 
$A\cap (Y\setminus Z)=\emptyset$, and $B\cap Z=\emptyset$.
\comment{
\begin{itemize}
\item  $A\cap Z\ne\emptyset$, 
\item  $B\cap (Y\setminus Z)\ne\emptyset$, 
\item  $A\cap (Y\setminus Z)=\emptyset$, and
\item  $B\cap Z=\emptyset$.
\end{itemize}
}
Note that when $\Pi$ is nondisjunctive,
this definition reduces to the corresponding definition given 
before. 

As with nondisjunctive programs, for any nonempty set~$Y$ of atoms 
that occur in $\Pi$, we say that $Y$ is {\sl elementary} for~$\Pi$ 
if all nonempty proper subsets of $Y$ are outbound in~$Y$ for $\Pi$. 
Similarly, every set consisting of a single atom occurring in $\Pi$ 
is an elementary set for~$\Pi$, and every elementary set for~$\Pi$ 
is a loop of $\Pi$. 
The definition of an elementary set for a disjunctive
program is stronger than the alternative definition of a loop 
provided in Proposition~\ref{prop:loop-alternative} for the
disjunctive case: It requires that the rules satisfy two additional
conditions, $A \cap (Y \setminus Z) = \emptyset$
and $B \cap Z = \emptyset$.

With these extended definitions, 
Propositions~\ref{prop:elm-def} and \ref{prop:elm-entail} remain 
correct after replacing ``nondisjunctive program'' in their 
statements with ``disjunctive program.''  
Theorem~\ref{thm:lf}(d) holds even when $\Pi$ is disjunctive.

To illustrate the definition, consider the following program:  \\[-1.2em]
\[ 
\ba l
  p\ ;\ q \ar p  \hspace{2.5em}
  p \ar q   \hspace{2.5em}
  p \ar \no\ r
\ea
\] 
Among the four loops of the program, $\{p\}$, $\{q\}$, $\{r\}$, and $\{p,q\}$,
the last one is not an elementary set because $\{q\}$ is not outbound 
in $\{p,q\}$: The first rule contains $q$ in the head and $p$ in 
the body, but it also contains 
\hbox{$\{p,q\}\cap (\{p,q\}\setminus \{q\}) = \{p\}$}
in the head.
According to the extension of Theorem~\ref{thm:lf}(d)
to disjunctive programs, the loop formula of $\{p,q\}$ can be 
disregarded.

\vspace{-0.2em}
\subsection{
Elementarily Unfounded Sets for Disjunctive Programs} 
\vspace{-0.2em}

Let $\Pi$ be a disjunctive program. 
For any sets $X$, $Y$ of atoms, 
by $\Pi_{X,Y}$ we denote the set of all rules~(\ref{abf-dis})
of $\Pi$ such that $X\sat B,F$ 
and $X\cap (A\setminus Y)=\emptyset$. 
Program $\Pi_{X,Y}$ contains all rules of~$\Pi$ that 
can provide supports for~$Y$ w.r.t.~$X$. 
If $\Pi$ is nondisjunctive and every atom $a_1$ in $Y$
has a rule~(\ref{abf-nd}) in~$\Pi$ such that $X\sat B,F$,
then $Y$ is elementary for~$\Pi_X$ iff it is elementary for~$\Pi_{X,Y}$.

\comment{
\optional{
\begin{prop} 
Let $\Pi$ be a nondisjunctive program, and $X$ a set of atoms, and 
$Y$ a set of atoms such that every element in $Y$ is externally 
supported by $\Pi$ w.r.t.~$X$.
Then $Y$ is an elementary set for~$\Pi_X$ iff it is an elementary 
set for~$\Pi_{X,Y}$.
\end{prop}
}
}

We extend the definition of an elementarily unfounded set to
disjunctive programs by replacing ``$\Pi_X$'' with ``$\Pi_{X,Y}$''
and by identifying $\Pi$ as a disjunctive program.
It is clear from the definition that every elementarily 
unfounded set for~$\Pi$ w.r.t.~$X$ is an elementary set for~$\Pi$ 
and that it is also an unfounded set for~$\Pi$ w.r.t.~$X$.

Propositions~\ref{prop:max-elm},~\ref{prop:min-uf},
Corollary~\ref{cor:elm-uf}, 
and Theorems~\ref{thm:lf}(e),~\ref{thm:lf}(e$'$) remain
correct after replacing 
``nondisjunctive program'' in their statements with ``disjunctive program''
and ``$\Pi_X$'' with ``$\Pi_{X,Y}$.''
For preserving the intended meaning of Theorem~\ref{thm:lf}(e),
``$Y$ is a maximal elementary set for~$\Pi_X$'' can be alternatively
replaced with ``$Y$ is maximal among all sets $Z$ of atoms that 
are elementary for $\Pi_{X,Z}$''

\vspace{-0.2em}
\subsection{Deciding Elementary Sets: Disjunctive Case}
\vspace{-0.2em}

Although deciding an elementary set can be done efficiently for
nondisjunctive programs, 
it turns out that the corresponding problem for 
(arbitrary) disjunctive programs is intractable.
\begin{prop}\label{theo:disj:coNP}
For any disjunctive program $\Pi$ and any set~$Y$ of atoms,
deciding whether~$Y$ is elementary for~$\Pi$ is {\sf coNP}-complete.
\end{prop}
This result can be explained by the close relationship to the problem
of deciding whether a set of atoms is {\sl unfounded-free}
\cite{lerusc97a}, which means that the set contains no nonempty 
unfounded subsets. 
In fact, the reduction from deciding unfounded-freeness to deciding
elementariness is straightforward. 

However, for the class of disjunctive programs called ``head-cycle-free''
\cite{bendec94a}, deciding an elementary set is tractable. 
A disjunctive program $\Pi$ is called \emph{head-cycle-free} 
if, for every rule~(\ref{abf-dis}) in $\Pi$, there is no loop $L$ of $\Pi$
such that $|A \cap L|> 1$. 
\comment{
For instance, the program 
\[ 
\ba l
p\,; q \ar  \\
p \ar q 
\ea
\] 
is head-cycle-free, while the program 
\[ 
\ba l
p\,; q \ar  \\
p \ar q \\
q \ar p
\ea
\] 
is not head-cycle-free.
}

The definition of an elementary subgraph for a nondisjunctive
program can be extended to a head-cycle-free program
by replacing
``(\ref{abf-nd})'' with ``(\ref{abf-dis})''
and ``$b\in B$'' with ``$a_1\in A,\ b\in B$''
in the equation for $\i{EC}_\Pi^{i+1}$.
\comment{
replacing the second equation by 
\begin{eqnarray*}
\i{EC}_\Pi^{i+1}(Y) &\hspace{-1em}=\hspace{-1em}& \i{EC}^i_\Pi(Y)\ \cup \\
   & &  \{ (a,b) \mid \text{there is a rule~(\ref{abf-dis}) in $\Pi$ 
such that } \\ 
   & &  a\in A,\; b\in B,\; \text{and the graph $(Y,\i{EC}_\Pi^i(Y))$ }  \\
   & &  \text{has a strongly connected subgraph containing} \\ 
   & &  \text{all atoms in $B\cap Y$} \} 
\end{eqnarray*}
}
With this extended definition of an elementary subgraph, 
Theorem~\ref{thm:ec-tr} remains correct after replacing 
``nondisjunctive program'' in its statement with 
``head-cycle-free program.''

\vspace{-0.7em}
\section{Comparison}
\vspace{-0.2em}
In this section, we compare our reformulation of elementary loops with
the original definition given in~\cite{gebsch05a} for nondisjunctive
programs. 

Let $\Pi$ be a nondisjunctive program. A loop of $\Pi$ is 
called {\sl trivial} if it consists of a single atom such that the
dependency graph of $\Pi$ does not contain an edge from the atom 
to itself. Non-trivial loops were called simply loops 
in~\cite{lin04,gebsch05a}. 
For a non-trivial loop~$L$, 
\[
   R^-_\Pi(L) = \{(\ref{abf-nd}) \in\Pi 
         \;\mid\; a_1\in L,\ \ B\cap L=\emptyset \}, 
\] \\[-1.9em]
\[
   R^+_\Pi(L) = \{(\ref{abf-nd}) \in\Pi 
         \;\mid\; a_1\in L,\ \ B\cap L\ne\emptyset \} .
\]

\begin{definition}~\cite[Definition~1]{gebsch05a} \label{def:elmloop}
Given a nondisjunctive program $\Pi$ and a non-trivial loop $L$ of $\Pi$, 
$L$ is called a {\sl GS-elementary loop} for~$\Pi$ if, for each non-trivial 
loop $L'$ of $\Pi$ such that $L'\subset L$, $R^-_\Pi(L')\cap 
R^+_\Pi(L)\ne\emptyset$.\footnote{A GS-elementary loop was called 
an ``elementary loop'' in \cite{gebsch05a}. Here we put ``GS-'' 
in the name, to distinguish it from a loop that is elementary 
under our definition.}
\end{definition}

\begin{prop}
For any nondisjunctive program $\Pi$ and any non-trivial loop $L$ of  $\Pi$,
$L$ is a GS-elementary loop for~$\Pi$ iff $L$ is an elementary set for $\Pi$.
\end{prop}

There are a few differences between Definition~\ref{def:elmloop}
and our definition of an elementary set. First, the definition of
an elementary set does not assume a priori that the set is a loop. 
Rather, the fact that an elementary set is a loop 
is a consequence of our definition. 
Second, our definition is simpler
because it does not refer to a dependency graph.
Third, the two definitions do not agree on trivial loops: A trivial
loop is an elementary set, but not a  GS-elementary loop. This originates
from the difference between the definition of a loop adopted 
in~\cite{lin04} and its reformulation given in~\cite{lee05}.
As shown in the main theorem of~\cite{lee05}, identifying a trivial
loop as a loop provides a simpler reformulation of the Lin-Zhao theorem 
by omitting reference to completion.
Furthermore, in the case of elementary sets, this reformulation also 
enables us to 
see a close relationship between maximal elementary sets (elementarily
unfounded sets)
and minimal nonempty unfounded sets. 
It also allows us to extend the notion of an elementary set 
to disjunctive programs without producing unintuitive results, 
unlike with GS-elementary loops. To see this, consider the following 
program: 
\beq
\ba l
 p\ ;\ q \ar r \hspace{2.5em}  
 p\ ;\ r \ar q \hspace{2.5em}
 q\ ;\ r \ar p \ . 
\ea
\eeq{ex-uf}
The non-trivial loops of this program are $\{p,q\}$, $\{p,r\}$, $\{q,r\}$,
and $\{p, q, r\}$, but not singletons $\{p\}$, $\{q\}$, and $\{r\}$.
If we were to extend GS-elementary loops to disjunctive programs, 
a reasonable extension would say that $\{p,q,r\}$ is a GS-elementary 
loop for program (\ref{ex-uf}) because all its non-trivial proper 
subloops are ``outbound'' in~$\{p,q,r\}$. 
Note that $\{p,q,r\}$ is unfounded w.r.t.~$\{p,q,r\}$. Moreover, 
every singleton is unfounded w.r.t~$\{p,q,r\}$ as well. This is 
in contrast with our Proposition~\ref{prop:max-elm}, 
according to which all nonempty 
proper subsets of an elementary set for program~(\ref{ex-uf}) 
w.r.t.~$\{p,q,r\}$ are externally supported w.r.t.~$\{p,q,r\}$.
This anomaly does not arise with our definition of an elementary set since
$\{p,q,r\}$ is not elementary for~(\ref{ex-uf}).
More generally, an elementary set is potentially elementarily
unfounded w.r.t.\ some model, which is not the case with
GS-elementary loops extended to disjunctive programs.

\vspace{-0.2em}
\section{Conclusion}

We have proposed the notion of an elementary set and provided 
a further refinement of the Lin-Zhao theorem based on it, which 
simplifies the Gebser-Schaub theorem and extends it to disjunctive 
programs. 

We have shown properties of elementary sets that allow us to 
disregard redundant loop formulas. 
One property is that, if all elementary subsets of a given set of atoms
are externally supported, the set is externally supported as well. 
Another property is that,
for a maximal set that is elementary for the relevant part of the
program w.r.t.\ some interpretation,
all its nonempty proper subsets are externally supported w.r.t.\ the same
interpretation.
Related to this, we have proposed the concept of elementarily unfounded sets,
which turn out to be precisely the minimal sets among nonempty 
unfounded sets.

Unlike elementary loops proposed in \cite{gebsch05a}, elementary
sets and the related results are extended to disjunctive 
programs in a straightforward way.
For nondisjunctive and head-cycle-free programs,
we have provided a graph-theoretic characterization of elementary sets,
which is simpler than the one proposed in~\cite{gebsch05a}.
For disjunctive programs,
we have shown that deciding elementariness is {\sf coNP}-complete,
which can be explained by the close relationship to deciding
unfounded-freeness of a given interpretation.

Elementary sets allow us to find more relevant unfounded sets
than what loops allow.
An apparent application is to consider elementarily 
unfounded sets in place of arbitrary unfounded loops as considered 
in the current SAT-based answer set solvers, at least for the tractable 
cases. For nondisjunctive programs,
an efficient algorithm for computing elementarily
unfounded sets is described in \cite{angesc06}, which 
can be extended to head-cycle-free programs as well.
Based on the theoretical foundations provided in this paper,
we plan to integrate elementarily unfounded set computation
into  {\sc cmodels} for an empirical evaluation. 

\vspace{-0.5em}
\section*{Acknowledgments}
We are grateful to Selim Erdo\u gan, Vladimir Lifschitz, Torsten Schaub,
and anonymous referees for their useful comments. Martin Gebser was
supported by DFG under grant SCHA 550/6-4, TP C. 
Yuliya Lierler was partially supported by the National
Science Foundation under Grant IIS-0412907.
\\[-2em]

\optional{
\section{Proofs} \label{sec:proofs}

\medskip
\noindent{\bf Proposition~\ref{prop:loop-alternative}$'$}\ \
{\it 
Let $\Pi$ be a disjunctive program. For any nonempty set $Y$ of atoms 
that occur in~$\Pi$, $Y$ is a loop of~$\Pi$ iff, for every nonempty 
proper subset $Z$ of $Y$, there is a rule~(\ref{abf-dis}) in~$\Pi$ 
such that 
\begin{itemize}
\item  $A\cap Z\ne\emptyset$, and
\item  $B\cap (Y\setminus Z)\ne\emptyset$.
\end{itemize} 
}\medskip

\optional{
\proof
{\sl Left to right:} Assume that $Y$ is a loop of $\Pi$. If $Y$ is 
a singleton, it is clear. If $Y$ is a non-singleton, take any nonempty
proper subset $Z$ of $Y$, and suppose, for the sake of contradiction, 
that there is no rule~(\ref{abf-dis}) in $\Pi$ such that 
$A\cap Z\ne\emptyset$, and $B\cap (Y\setminus Z)\ne\emptyset$. 
It follows that there is no path from an atom 
in $Z$ to an atom in $Y\setminus Z$, which contradicts that $Y$ is a loop 
of $\Pi$.

{\sl Right to Left:} Assume that $Y$ is not a loop of $\Pi$. Take any atom 
$p$ in $Y$, and let $Z$ be the set of atoms that are reachable from $p$ in 
the dependency graph of $\Pi$. Since $Y$ is not a loop, 
$Y\setminus Z\ne\emptyset$. Since there is no edge from an atom in $Z$
to an atom in $Y\setminus Z$ in the dependency graph, it follows that 
there is no rule~(\ref{abf-dis}) in $\Pi$ such that 
$A\cap Z\ne\emptyset$ and $B\cap (Y\setminus Z)\ne\emptyset$.
\qed
}

\medskip\noindent
{\bf Proposition~\ref{prop:elm-def}$'$}
{\it 
For any disjunctive program~$\Pi$ and for any nonempty set $Y$ 
of atoms that occur in~$\Pi$, $Y$ is an elementary set for~$\Pi$ 
iff all loops $Z$ of $\Pi$ such that $Z\subset Y$ are outbound 
in $Y$ for $\Pi$.
}\medskip


\proof
 From left to right is clear. 

{\sl Right to left:}
Assume that every loop is outbound in $Y$ for $\Pi$.
Let $Z$ be a nonempty subset of $Y$. The subgraph of the dependency graph 
of $\Pi$ induced by $Z$ has a strongly connected component $Z'$ 
that has no outgoing edges. 

Since $Z'$ is a loop of $\Pi$ that is contained in $Y$, by the assumption that
$Z'$ is outbound in $Y$ for $\Pi$,  there is a rule~(\ref{abf-dis}) 
such that 
\beq  A\cap Z' \ne\emptyset, \eeq{e1}
\beq  A\cap (Y\setminus Z') = \emptyset, \eeq{e2}
\beq  B\cap Z'=\emptyset, \eeq{e3}
and
\beq  B\cap (Y\setminus Z')\ne\emptyset. \eeq{e4}

For this rule, since $Z'\subseteq Z$, it follows from~(\ref{e1}), (\ref{e2}) 
that 
\beq 
   A\cap Z\ne\emptyset, 
\eeq{z1} 
and 
\beq 
   A\cap (Y\setminus Z)=\emptyset. 
\eeq{z2}
Furthermore, it follows that $B\cap (Z\setminus Z')=\emptyset$ 
from the assumption that $Z'$ has no outgoing edges in the subgraph. 
This, in combination with~(\ref{e3}), (\ref{e4}), yields
\beq
   B\cap Z=\emptyset, 
\eeq{z3}
and
\beq
   B\cap (Y\setminus Z)\ne\emptyset. 
\eeq{z4} 
  From~(\ref{z1}), (\ref{z2}), (\ref{z3}), (\ref{z4}), we conclude
that $Z$ is outbound in $Y$ for $\Pi$.
\qed

\medskip\noindent
{\bf Proposition~\ref{prop:elm-entail}$'$}
{\it 
Let $\Pi$ be a disjunctive program, $X$ a set of atoms, and 
$Y$ a nonempty sets of atoms.
If $X\sat\i{LF}_\Pi(Z)$ for all elementary sets $Z$ for $\Pi$
that are contained in $Y$, then $X\sat\i{LF}_\Pi(Y)$.
}\medskip

\comment{
\medskip\noindent
{\bf Proposition~\ref{prop:elm-entail}$'$}
{\it 
Let $\Pi$ be a disjunctive program, $X$ a set of atoms, and 
$Y$ a subset of $X$.
If $X\sat\i{ES}_\Pi(Z)$ for all elementary sets $Z$ for $\Pi$
that are contained in $Y$, then $X\sat\i{ES}_\Pi(Y)$.
}\medskip
}


The proposition follows from the following lemma and the fact that
there is a maximal set among the elementary sets for $\Pi$ that is 
contained in $Y$ and is not outbound in~$Y$. 

\begin{lemma} \label{lem:nob-entail}
Let $\Pi$ be a disjunctive program, $X$ a set of atoms, 
$Y$ a subset of $X$, and $Z$ a nonempty subset of $Y$.
If $Z$ is not outbound in $Y$ and $X\sat\i{ES}_\Pi(Z)$, 
then $X\sat\i{ES}_\Pi(Y)$.
\end{lemma}

\proof
Assume that $Z$ is not outbound in $Y$, and that $X\sat\i{ES}_\Pi(Z)$.
There is a rule~(\ref{abf-dis}) in $\Pi$ such that 
\beq 
  A\cap Z\ne\emptyset ,
\eeq{az}
\beq
  B\cap Z=\emptyset ,
\eeq{bz}
\beq
  X\sat B, F ,
\eeq{xbf}
and
\beq
  X\cap (A\setminus Z)=\emptyset .
\eeq{xaz}
 From~(\ref{az}), since $Z\subseteq Y$, 
\beq
  A\cap Y\ne\emptyset, 
\eeq{ay}
 From the fact that $Y\subseteq X$ and (\ref{xaz}), 
$$
  Y\cap(A\setminus Z) = \emptyset, 
$$
which is equivalent to
\beq
  A\cap (Y\setminus Z) = \emptyset .
\eeq{ayz}
 Since $Z$ is not outbound in $Y$, from~(\ref{az}), (\ref{bz}), (\ref{ayz}), 
 it follows that 
$$ 
 B\cap (Y\setminus Z) = \emptyset, 
$$ 
which, in combination with~(\ref{bz}), gives us that
\beq
 B\cap Y=\emptyset. 
\eeq{by}

 From~(\ref{xbf}), (\ref{ay}) and (\ref{by}), we conclude that 
$X\sat\i{ES}_\Pi(Y)$.
\qed

\comment{
Assume that $Z$ is not outbound in $Y$, that $X\sat\i{LF}_\Pi(Z)$, and 
that $X\sat\bigwedge Y$, i.e.,
\beq
  Y\subseteq X.
\eeq{yx} 
Since $Z\subseteq Y$, it follows from~(\ref{yx}) that $X\sat\bigwedge Z$, 
and by the assumption that $X\sat\i{LF}_\Pi(Z)$, it follows that 
$X\sat\i{ES}_\Pi(Z)$, i.e., there is a rule~(\ref{abf-dis}) 
in $\Pi$ such that 
}

\begin{lemma} \label{lem:elm-x}
Let $\Pi$ be a disjunctive program, and $X$ a model of $\Pi$, 
and $Y$ a set of atoms. Every elementary set for~$\Pi_{X,Y}$ belongs to $X$.
\end{lemma}

\medskip\noindent
{\bf Proposition~\ref{prop:max-elm}$'$} 
{\it 
For any disjunctive program~$\Pi$, any set $X$ of atoms, and 
any elementary set $Y$ of $\Pi_{X,Y}$, $X\sat\i{LF}_\Pi(Z)$
for all nonempty proper subsets $Z$ of $Y$.
}\medskip

\proof
\comment{
Since $Y$ is an elementary set of $\Pi_{X,Y}$, for every nonempty subset 
$Z$ of $Y$ there is a rule~(\ref{abf-dis}) such that 
$Y\setminus Z \subseteq B\subseteq X$. It follows that $Y\subseteq X$.
By the assumption $X\sat\i{ES}_\Pi(Y)$, so that there is a
 rule~(\ref{abf-dis}) in $\Pi_{X,Y}$ such that 
$A\in Y$, $B\cap Y=\emptyset$, $X\sat B,F$, and 
$X\cap (A\setminus Y)=\emptyset$.
-------
}
Since $Z$ is outbound in $Y$ for $\Pi_{X,Y}$, there is a 
rule~(\ref{abf-dis}) in $\Pi$ such that 
\beq 
  A\cap Z\ne\emptyset,
\eeq{m-az} 
\beq 
  B\cap Z=\emptyset,
\eeq{m-bz}
\beq
  A\cap (Y\setminus Z)=\emptyset.
\eeq{m-ayz}
\beq
  X\sat B,F, 
\eeq{m-xbf}
and 
\beq 
   X\cap (A\setminus Y)=\emptyset. 
\eeq{m-xay}
 From~(\ref{m-ayz}) and (\ref{m-xay}), it follows that 
\beq
  X\cap (A\setminus Z)=\emptyset.
\eeq{m-xaz}

 From~(\ref{m-az}), (\ref{m-bz}), (\ref{m-xbf}), (\ref{m-xaz})
we conclude that $X\sat\i{ES}_\Pi(Z)$.
\qed

\medskip
\noindent{\bf Theorem~\ref{thm:lf}(e)}\ \
{\it 
The following condition is equivalent to conditions~(a)--(c) of 
Theorem~\ref{thm:lf}$'$.
\begin{itemize}
\item[(e)]  $X$ satisfies $\i{LF}_\Pi(Y)$ for every $Y$ such that 
             \begin{itemize}
            \item  $Y$ is a maximal elementary set of $\Pi_{X,Y}$, or
            \item  $Y$ is a singleton set of an atom occurring in $\Pi$.
            \end{itemize}
\end{itemize}
}


\proof
We will prove the equivalence between (b) and (e) of Theorem~\ref{thm:lf}:  
 From (b) to (e) is clear.

{\sl From (e) to (b):} Assume (e). Let $Y$ be a subset of $X$.
Since every atom in $Y$ is externally supported w.r.t.~$X$, 
it follows that there is a maximal elementary set $Z$ in $\Pi_{X,Y}$
that is externally supported. By Lemma~\ref{lem:elm-x}, $Z\subseteq X$. 
Since $\Pi_{X,Y}\subseteq \Pi_{X,Z}$, $Z$ is an elementary set of \
$\Pi_{X,Z}$, and since every elementary set of $\Pi_{X,Z}$ belongs to $X$
(by Lemma~\ref{lem:elm-x}) and $X\cap(A\setminus Z)=\emptyset$ for all 
rules~(\ref{abf-dis}) in $\Pi_{X,Z}$, it follows that $Z$ is a maximal
elementary set of $\Pi_{X,Z}$.  By assumption, $Z$ is externally supported
w.r.t.~$X$. By Proposition~\ref{prop:max-elm}, 
it follows that every elementary set in $Y$ is externally supported 
w.r.t.~$X$. Then by Proposition~\ref{prop:elm-entail}, it follows 
that $Y$ is externally supported by w.r.t.~$X$.
\qed

\comment{
$X\cap (A\setminus Z)=\emptyset$  check that
$Z$ is a maximal elementary set of $\Pi_{X,Z}$. Note that $X\cap 

[[ For every singeton in $X$, $X\sat\i{ES}(\Pi)$. 
   
   For every $Y$, $X\sat\

all maximal elementary
sets $Y$ of $\Pi_{X,Y}$ is externally supported by $\Pi$ w.r.t.~$X$, 
by Proposition~\ref{prop:max-elm}, all elementary sets $Y$ for $\Pi_{X,Y}$
is externally supported by $\Pi$ w.r.t.~$X$. For an elementary set $Z$
of $\Pi$ that is not an elementary set of $\Pi_{X,Y}$, it follows that 
$Z\not\subseteq X$. Indeed, otherwise, $X\sat\i{ES}_\Pi(Z)$, but this 
contradicts 
but not

[[ every non-singleton elementary set $Z$ of $\Pi_{X,Y}$ 
   is that $Z\subseteq X$ ]]

$X$ satisfies $\i{ES}_\Pi(Y)$
for all maximal elementary sets $Y$ for $\Pi_{X,Y}$. 
Since $X$ is a model of $\Pi$, it follows that $Y\subseteq X$. 
By the assumption
that every maximal elementary set for $\Pi_{X,Y}$ is externally 
supported w.r.t.~$X$, and by Proposition~\ref{prop:max-elm}, 
it follows that all elementary sets for $\Pi_{X,Y}$ are externally 
supported w.r.t.~$X$. By Proposition~\ref{prop:elm-entail}, 
it follows that $Y$ is externally supported w.r.t.~$X$.
\qed
}

\medskip
\noindent
{\bf Corollary~\ref{cor:elm-uf-max}$'$}
{\it 
For any disjunctive program $\Pi$ and any set $X$ of atoms, 
every elementarily unfounded set for~$\Pi$ w.r.t.~$X$ is a maximal 
elementary set of $\Pi_{X,Y}$.
}\medskip

[[ JL: need to change the statement ]]

\proof
Let $Y$ be an elementarily unfounded set of $\Pi$ w.r.t.~$X$, but 
for the sake of contradiction that it is not maximal elementary set of
$\Pi_{X,Y}$. Then there is another elementarily unfounded set $Z$ of $\Pi$
w.r.t.~$X$ such that $Y\subset Z$. 
By Proposition~\ref{prop:max-elm}, $X$ satisfies $\i{LF}_\Pi(Y)$. 
Since $Y$ is an elementary set of $\Pi_X$, 
it follows that $Y\subseteq X$, so that $X\sat\i{ES}_\Pi(Y)$, which 
contradicts that $Y$ is  unfounded w.r.t.~$X$.
\qed

\comment{
\medskip\noindent
{\bf Proposition~\ref{prop:elm-uf}$'$} 
{\it 
Let $\Pi$ be a disjunctive program, $X$ a set of atoms, and $Y$ 
an elementarily unfounded set of $\Pi$ w.r.t.~$X$. Then
\begin{itemize}
\item[(i)]   $X$ does not satisfy $\i{LF}_\Pi(Y)$, and
\item[(ii)]  $X$ satisfies $\i{LF}_\Pi(Z)$ for all nonempty subsets $Z$ of $Y$.
\end{itemize}
}\medskip

\proof
First we will check that condition~(i) holds.
Since $Y$ is an elementary set of $\Pi_{X,Y}$, for every nonempty subset 
$Z$ of $Y$ there is a rule~(\ref{abf-dis}) such that 
$Y\setminus Z \subseteq B\subseteq X$. It follows that $Y\subseteq X$.
On the other hand, one can check that $Y$ is unfounded by $\Pi$ w.r.t.~$X$: 
it is clear that $Y$ is unfounded by $\Pi\setminus\Pi_X$ 
w.r.t.~$X$, and that $Y$ is unfounded by $\Pi_X$ w.r.t.~$X$. 
Thus we conclude that condition (i) holds.

Now we will check that condition (ii) holds.
Since $Y$ is elementary in~$\Pi_{X,Y}$, 
for any nonempty subset $Z$ of $Y$, there is a rule~(\ref{abf-dis}) 
in $\Pi$ such that 
\beq  A\cap Z\ne\emptyset, \eeq{i}
\beq  A\cap (Y\setminus Z)=\emptyset, \eeq{ii}
\beq  B\cap Z=\emptyset, \eeq{iii}
$$ 
    B\cap (Y\setminus Z)\ne\emptyset, 
$$ 
\beq  X\sat B, F, \eeq{v} 
and
\beq  X\cap (A\setminus Y)=\emptyset.  \eeq{vi}

 From~(\ref{ii}) and (\ref{vi}), it follows that 
\beq 
  X\cap (A\setminus Z)=\emptyset.
\eeq{vii} 
 From~(\ref{i}), (\ref{iii}), (\ref{v}), (\ref{vii}), we conclude
that $Z$ is externally supported by $\Pi$ w.r.t.~$X$.
\qed
}

\medskip\noindent
{\it
{\bf Proposition~\ref{prop:max-min}$'$}
For any disjunctive program $\Pi$ and any sets $X$, $Y$ of atoms, 
$Y$ is a (maximal) elementary set for $\Pi_{X,Y}$ that is unfounded 
by $\Pi$ w.r.t.~$X$ iff $Y$ is minimal among the nonempty sets of atoms
occurring in $\Pi_{X,Y}$ that are unfounded by $\Pi$ w.r.t.~$X$.
}\medskip

\proof
 From left to right follows from Corollary~\ref{cor:elm-uf} (by 
identifying $\Pi$ as a disjunctive program).

{\sl Right to left:}
Assume that $Y$ is minimal among the nonempty unfounded sets for~$\Pi_{X,Y}$
whose atoms occur in $\Pi_{X,Y}$.

Case 1: $Y$ is a singleton. It is clearly an elementary
set of $\Pi_{X,Y}$ that is unfounded w.r.t.~$X$. By 
Proposition~\ref{prop:max-elm}, it follows that it is maximal.

Case 2: otherwise. From the minimality assumption on $Y$, it follows
that $Y\subseteq X$. Indeed, if $Y\not\subseteq X$, then there is 
an atom $a\in Y\setminus X$, and one can check that $a$ is also 
unfounded by $\Pi_{X,Y}$ w.r.t.~$X$, which contradicts that $Y$ is minimal.
It is also that $X\sat\i{ES}_\Pi(Z)$ for 
every nonempty proper subset $Z$ of $Y$, so that there is 
a rule~(\ref{abf-dis}) in $\Pi$ such that 
\beq  
   A\cap Z\ne\emptyset, 
\eeq{mm-az}
\beq  
   B\cap Z=\emptyset, 
\eeq{mm-bz}
and
\beq  
   X\sat B, F, and
\eeq{mm-xbf} 
\beq  
   X\cap (A\setminus Z)=\emptyset.
\eeq{mm-xaz}
 From~(\ref{mm-az}) and the fact that $Z\subset Y$, it follows that
\beq
   A\cap Y\ne\emptyset, 
\eeq{mm-ay}
and
\beq  
   X\cap (A\setminus Y)=\emptyset.
\eeq{mm-xay}
Since $Y$ is unfounded by $\Pi_X$, it follows that 
\beq
   B\cap Y\ne\emptyset.
\eeq{mm-by}
 From~(\ref{mm-xaz}), (\ref{mm-xay}), and the fact that 
$Z\subset Y\subseteq X$, it follows that 
\beq
   A\cap (Y\setminus Z)=\emptyset.
\eeq{mm-ayz}
 From~(\ref{mm-az}), (\ref{mm-bz}), (\ref{mm-by}), (\ref{mm-ayz}), 
we conclude that $Y$ is elementary in $\Pi_X$. Since $Y$ is unfounded
by $\Pi_X$, clearly it is also unfounded by $\Pi$. 
Therefore, $Y$ is elementarily unfounded set of $\Pi$.
\qed

\medskip\noindent
{\bf Proposition~\ref{prop:min-uf}$'$}
{\it 
For any disjunctive program $\Pi$ and any sets $X$, $Y$ of atoms, 
$Y$ is an elementarily unfounded set of $\Pi$ w.r.t.~$X$ iff 
$Y$ is minimal among the nonempty sets of atoms occurring in $\Pi$
that are unfounded by $\Pi$ w.r.t.~$X$.
}\medskip


\proof
 From left to right follows from Corollary~\ref{cor:elm-uf} (by 
identifying $\Pi$ as a disjunctive program). 

{\sl Right to left:}
Assume that $Y$ is minimal among the nonempty unfounded sets for~$\Pi$
w.r.t.~$X$ whose atoms occur in $\Pi$.

Case 1: $Y$ is a singleton. It is clearly an elementarily 
unfounded set of $\Pi$ w.r.t.~$X$.

Case 2: otherwise. From the minimality assumption on $Y$, it follows
that $Y\subseteq X$. Indeed, if $Y\not\subseteq X$, then there is 
an atom $a\in Y\setminus X$, and one can check that $a$ is also 
unfounded by $\Pi_{X,Y}$ w.r.t.~$X$, which contradicts that $Y$ is minimal.
It is also that $X\sat\i{ES}_\Pi(Z)$ for
every nonempty proper subset $Z$ of $Y$, so that there is 
a rule~(\ref{abf-dis}) in $\Pi$ such that 
\beq  
   A\cap Z\ne\emptyset, 
\eeq{u-az}
\beq  
   B\cap Z=\emptyset, 
\eeq{u-bz}
and
\beq  
   X\sat B, F, 
\eeq{u-xbf} 
and
\beq  
   X\cap (A\setminus Z)=\emptyset.
\eeq{u-xaz}
 From~(\ref{u-az}) and the fact that $Z\subset Y$, it follows that
\beq
   A\cap Y\ne\emptyset, 
\eeq{u-ay}
and
\beq  
   X\cap (A\setminus Y)=\emptyset.
\eeq{u-xay}
Since $Y$ is unfounded by $\Pi_X$, it follows that 
\beq
   B\cap Y\ne\emptyset.
\eeq{u-by}
 From~(\ref{u-xaz}), (\ref{u-xay}), and the fact that 
$Z\subset Y\subseteq X$, it follows that 
\beq
   A\cap (Y\setminus Z)=\emptyset.
\eeq{u-ayz}
 From~(\ref{u-az}), (\ref{u-bz}), (\ref{u-by}), (\ref{u-ayz}), 
we conclude that $Y$ is elementary in $\Pi_X$. Since $Y$ is unfounded
by $\Pi_X$, clearly it is also unfounded by $\Pi$ w.r.t.~$X$.
Therefore, $Y$ is elementarily unfounded set of $\Pi$.
\qed

\comment{
For a minimal unfounded set~$Y$ w.r.t.~$X$ that is no singleton,\footnote{%
If $Y$ is a singleton, it is an elementarily unfounded set w.r.t.~$X$.}
we have $Y \subseteq X$ and $X \sat \i{ES}_\Pi(Z)$ 
for every nonempty proper subset~$Z$ of~$Y$.\footnote{%
For any unfounded set $Y \not\subseteq X$ w.r.t.~$X$ such that~$Y$ is 
no singleton, $Y \setminus \{a\}$  for any $a \in Y \setminus X$
is as well an unfounded set w.r.t.~$X$; therefore, such an unfounded set~$Y$
is not minimal w.r.t.~$X$.}
This implies that, for each such subset $Z$, there is a rule~(\ref{abf-dis})
in~$\Pi$ such that 
\beq  A\cap Z\ne\emptyset, \eeq{uf1}
\beq  A\cap (X\setminus Z)=\emptyset, \eeq{uf2}
\beq  B\cap Z=\emptyset, \eeq{uf3}
and
\beq  X\sat B, F. \eeq{uf4} 
Since $Y \subseteq X$, any rule~(\ref{abf-dis}) that satisfies~(\ref{uf2})
also satisfies $A\cap (Y\setminus Z) = \emptyset$.
Because~$Y$ is unfounded, $B \cap Y \ne\emptyset$ holds for every rule 
satisfying~~(\ref{uf1}),~(\ref{uf2}),~(\ref{uf3}), and~(\ref{uf4}).
Hence, $Z$ is outbound in~$Y$ for~$\Pi_{X,Y}$.
As~$Z$ is an arbitrary nonempty proper subset of $Y$,
it follows that~$Y$ is an elementary set of $\Pi_{X,Y}$
and thus an elementarily unfounded set w.r.t.~$X$.
\qed
}

\comment{
\begin{lemma}
Let $\Pi$ be a disjunctive program, $Y$ a set of atoms occurring in $\Pi$,
and $Z$ a nonempty proper subset of $Y$. Then,
$Z$ is outbound in $Y$ for $\Pi$ iff there is an edge from an atom in $Z$ 
to an atom in $Y\setminus Z$ in the elementary subgraph of $Y$ for $\Pi$.
\end{lemma}

\proof
{\sl From right to left:}
Assume $Z$ is not outbound in $Y$ for $\Pi$. 
We will show that the graph $(Y, \i{EC}_\Pi^i(Y))$ $(i\ge 0)$ has 
no edges from an atom in $Z$ to an atom in $Y\setminus Z$. 
First it's clear that graph $(Y,\i{EC}_\Pi^0(L))$ has no such edge. 
Assume that $(Y, \i{EC}_\Pi^i(Y))$ has no such edge. 
Consider any rule~(\ref{abf-nd}) such that $a_1\in Z$.  

Case 1: $B\cap (Y\setminus Z)=\emptyset$. Clearly the rule cannot contribute
to the edges from $Z$ to $Y\setminus Z$ in $(Y, \i{EC}_\Pi^{i+1}(Y))$.

Case 2: $B\cap (Y\setminus Z)\ne\emptyset$. Since $Z$ is not outbound 
        in $Y$, it follows that $B\cap Z\ne\emptyset$. Since there is no 
        loop that contains $(B\cap Z)\cup (B\cap (Y\setminus Z))= B\cap Y$, 
        the rule cannot contribute to the edges from $Z$ to $Y\setminus Z$
        in $(Y,\i{EC}_\Pi^{i+1}(Y))$. 

Consequently, $(Y,\i{EC}_\Pi^{i+1}(Y))$ has no edges from $Z$ to 
$Y\setminus Z$.

\medskip

{\sl From left to right:}

\qed
}

\comment{
The following lemma tells the relationship between elementary sets 
and elementary subgraphs.
\begin{lemma} \label{lem:ec-tr}
For any nondisjunctive program $\Pi$ and any set $Y$ of atoms
occurring in $\Pi$, 
$Y$ is an elementary set of $\Pi$ iff, for any nonempty proper subsets 
$Z$ of $Y$, the elementary subgraph of $Y$ in $\Pi$ has an edge from 
an atom in $Z$ to an atom in $Y\setminus Z$.
\end{lemma}

\proof
{\sl Right to left:} Assume $Y$ is not elementary. Then there is a proper
subset $Z$ of $Y$ such that $Z$ is not outbound in $Y$. 
We will show that graph $(Y, \i{EC}_\Pi(Y))$ has no edges from an atom in 
$Z$ to an atom in $Y\setminus Z$. 
First it's clear that graph $(Y,\i{EC}_\Pi^0(L))$ has no such edge. 
Assume that $(Y, \i{EC}_\Pi^i(Y))$ has no such edge. 
Consider any rule~(\ref{abf-nd}) such that $a_1\in Z$.  

Case 1: $B\cap (Y\setminus Z)=\emptyset$. Clearly the rule cannot contribute
to the edges from $Z$ to $Y\setminus Z$ in $(Y, \i{EC}_\Pi^{i+1}(Y))$.

Case 2: $B\cap (Y\setminus Z)\ne\emptyset$. Since $Z$ is not outbound 
        in $Y$, it follows that $B\cap Z\ne\emptyset$. Since there is no 
        loop that contains $(B\cap Z)\cup (B\cap (Y\setminus Z))= B\cap Y$, 
        the rule cannot contribute to the edges from $Z$ to $Y\setminus Z$
        in $(Y,\i{EC}_\Pi^{i+1}(Y))$. 

Consequently, $(Y,\i{EC}_\Pi^{i+1}(Y))$ has no edges from $Z$ to 
$Y\setminus Z$.

{\sl Left to right:} Assume that $\i{EC}$ is not strongly connected. Then 
there is a strongly connected component $C$ such that $C$ has no incoming
edges in $\i{EC}$. Consider the subgraph of the dependency graph of 
$\i{EC}$ induced by $L\setminus C$. There exists a strongly connected 
component $D$ in that subgraph that has no outgoing edges. 
Consider any rule $a\ar B,F$ such that $a\in D$, and $B\cap D=\emptyset$.
By the choice of $D$, $B$ does not contain an atom in $L\setminus (C\cup D)$. 
There is a strongly connected component in $\i{EC}$ that contains $a$. 
We can check that $B$ contains no atom in $C$. Indeed if $B$ contains
an atom in $C$, then $C$ must have an incoming edge in $\i{EC}$, which 
contradicts the choice of $C$.
Therefore $B$ does not contain an atom in 
$$ C\cup (L\setminus (C\cup D) \cup D) = L. $$
Therefore the rule belongs to $R^-(L)$, which means that $L$ is not 
elementary.

\qed
Note that the left-to-right direction also holds when ``elementary subgraph'' 
in the statement is replaced with ``dependency graph,''  
but the right-to-left direction does not. This difference allows us to 
use the elementary subgraph to decide elementary sets.

\proof
{\sl Right to left:} 

\proof
{\sl From right to left:}

{\sl Left to right:}

\qed
}

\noindent{\bf Theorem~\ref{thm:ec-tr}$'$}\ \
{\it 
For any head-cycle-free program $\Pi$ and any set $Y$ of atoms 
occurring in $\Pi$, $Y$ is an elementary set of $\Pi$ iff 
the elementary subgraph of $Y$ for $\Pi$ is strongly connected.
}\medskip

\proof
{\sl Left to right:} 
Assume that the elementary subgraph of~$Y$ for~$\Pi$ is not 
strongly connected. There is a strongly connected component in
the graph that has no incoming edges. Let the set of atoms in that
component be $C$. 
Consider any rule (\ref{abf-dis}) in $\Pi$ such that
$A\cap (Y\setminus C)\ne\emptyset$ and $B\cap C\ne\emptyset$. 
Since the elementary subgraph has no edge from an atom 
in $Y\setminus C$ to $C$, 
it follows that $B\cap (Y\setminus C)\ne\emptyset$. 
\comment{
Case 1: $Y\setminus C$ contains to a nontrivial loop:
        Since $\Pi$ is head-cycle-free, it follows that 
        $A\cap C=\emptyset$.

Case 2: $Y\setminus C$ does not contain a nontrivial loop:
}
Therefore, $Y\setminus C$ is not outbound in $Y$ for $\Pi$,
so that $Y$ is not an elementary set of $\Pi$.
\medskip

{\sl Right to left:} 
Assume that $Y$ is not an elementary set of $\Pi$. Then there is a 
loop $Z\subset Y$ which is not outbound in $Y$ for $\Pi$. 
We will show that the graphs $(Y, \i{EC}_\Pi^i(Y))$ $(i\ge 0)$ has 
no edges from an atom in $Z$ to an atom in $Y\setminus Z$. 
First, it's clear that graph $(Y,\i{EC}_\Pi^0(L))$ has no such edge. 
Next, assume that $(Y, \i{EC}_\Pi^i(Y))$ has no such edge. 
Consider any rule~(\ref{abf-dis}) such that $A\cap Z\ne\emptyset$, 
$B\cap (Y\setminus Z)\ne\emptyset$.  Since $\Pi$ is head-cycle-free,
it follows that $A\cap (Y\setminus Z)=\emptyset$. From the assumption
that $Z$ is $Z$ is not outbound in $Y$, it follows that 
$B\cap Z\ne\emptyset$. Since there is no 
        cycle  in $(Y,EC^i_\Pi(Y))$ that contains 
        $(B\cap Z)\cup (B\cap (Y\setminus Z))= B\cap Y$, 
        the rule cannot contribute to the edges from $Z$ to $Y\setminus Z$
        in $(Y,\i{EC}_\Pi^{i+1}(Y))$. 
Consequently, $(Y,\i{EC}_\Pi^{i+1}(Y))$ has no edges from $Z$ to 
$Y\setminus Z$. 
\qed

\comment{
\proof
We show 
(1) that $Y$ is an elementary set of $\Pi$ 
if its elementary subgraph is strongly connected and
(2) that $Y$ is no elementary set of $\Pi$ otherwise.
\begin{itemize}
\item[(1)] 
Let the elementary subgraph of~$Y$ in~$\Pi$ be strongly connected.
For any nonempty proper subset~$Z$ of~$Y$,
there is at least one edge from an atom in~$Z$ to an atom in~$Y \setminus Z$.
By the construction of the elementary subgraph, 
there is a rule~(\ref{abf-dis}) in~$\Pi$ such that
\beq  A\cap Z \ne\emptyset,\footnote{%
Note that~(\ref{hcf1}) implies $A\cap (Y\setminus Z) = \emptyset$
because~$Y$ is a loop and~$\Pi$ is head-cycle-free.} \eeq{hcf1}
\beq  B\cap Z = \emptyset, \eeq{hcf2}
and
\beq  B\cap (Y\setminus Z) \ne\emptyset.\footnote{%
If no rule~(\ref{abf-dis}) in~$\Pi$, for which~(\ref{hcf1}) 
and~(\ref{hcf3}) hold, satisfies~(\ref{hcf2}),
an atom in~$Z$ cannot have an edge to any atom in~$Y \setminus Z$
because the atoms of $B \cap Y$ are then not contained in a 
strongly connected subgraph.} \eeq{hcf3}
That is, $Z$ is outbound in $Y$ for $\Pi$.
Since $Z$ is an arbitrary nonempty proper subset of $Y$,
it follows that $Y$ is an elementary set.
\item[(2)]
Let the elementary subgraph of~$Y$ in~$\Pi$ be not strongly connected.
Then, there is a strongly connected~$C$ such that no atom
in $Y \setminus C$ has an edge to any atom in~$C$.
By the construction of the elementary subgraph,
we have $B\cap (Y\setminus C) \ne\emptyset$ for any
rule~(\ref{abf-dis}) in $\Pi$ such that 
$A\cap (Y\setminus C) \ne\emptyset$ and $B\cap C \ne\emptyset$.
Hence, $Y\setminus C$ is not outbound in $Y$ for $\Pi$,
and $Y$ is no elementary set of $\Pi$.
\qed
\end{itemize}
}


\medskip
\noindent{\bf Proposition~\ref{theo:disj:coNP}} \ \
{\it 
For any disjunctive program $\Pi$, deciding whether a set $Y$ of 
atoms is elementary for $\Pi$ is {\sf coNP}-complete.
}\medskip

\proof
We first show membership in {\sf coNP} and afterwards 
{\sf coNP}-hardness of deciding whether~$Y$ 
is an elementary set of~$\Pi$.

Consider the complementary problem to decide whether~$Y$ is no
elementary set of~$\Pi$.
We can nondeterministically guess a nonempty 
proper subset~$Z$ of~$Y$.
Verifying that for every rule~(\ref{abf-dis}) in~$\Pi$ either
\begin{itemize}
\item $A\cap Z = \emptyset$,
\item $A\cap (Y\setminus Z) \ne\emptyset$,
\item $B\cap Z \ne\emptyset$, or
\item $B\cap (Y\setminus Z) = \emptyset$
\end{itemize}
is straightforward.
Hence the complementary problem is in {\sf NP},
the original problem is in {\sf coNP}.

Next we show {\sf coNP}-hardness by reducing 3-UNSAT.
We represent a propositional formula~$F$ in 3-CNF
as set of clauses
$F = \{l^1_1 \vee l^2_1 \vee l^3_1\,,\, \ldots\,,\, l^1_m \vee l^2_m \vee l^3_m\}$.
We let $\{x_1,\ldots,x_n\}$ be
the set of propositional variables occurring in~$F$.
A propositional variable~$x$ is associated with positive literal~$x$
and negative literal~$\overline{x}$.
For a literal~$x$ or~$\overline{x}$,
we define the complement~$\overline{x}$ or~$\overline{\overline{x}}$
as~$\overline{x}$ or~$x$, respectively.
By $Y = \{x_1,\overline{x_1},\ldots,x_n,\overline{x_n}\}$, we denote the
set of literals associated with $F$.

In what follows, we reduce~$F$ to a disjunctive program~$\Pi_F$
such that~$Y$ is an elementary set of~$\Pi_F$ iff~$F$ is unsatisfiable.
The reduction consists of three parts (subprograms).
For a nonempty proper subset~$Z$ of~$Y$, the three parts guarantee the
following:
\begin{enumerate}
\item $Z$ is outbound in~$Y$ for~$\Pi_F$ if it contains neither~$x$ 
      nor~$\overline{x}$ for some variable~$x$.
\item $Z$ is outbound in~$Y$ for~$\Pi_F$ if it contains 
      both~$x$ and~$\overline{x}$ for some variable~$x$.
\item $Z$ is outbound in~$Y$ for~$\Pi_F$ if some clause in~$F$
      is not satisfied by the literals in~$Z$.
\end{enumerate}

The following subprogram constitutes the first part of the reduction:
\beq
\Pi_z =
\left\{
\begin{array}{l@{\hspace{15mm}}l}
x_1 \ar x_2,\overline{x_2}
& 
\overline{x_1} \ar x_2,\overline{x_2}
\\
x_2 \ar x_3,\overline{x_3}
&
\overline{x_2} \ar x_3,\overline{x_3}
\\[0mm]
\;\vdots
&
\;\vdots
\\[0mm]
x_{n-1} \ar x_n,\overline{x_n}
&
\overline{x_{n-1}} \ar x_n,\overline{x_n}
\\
x_n \ar x_1,\overline{x_1}
&
\overline{x_n} \ar x_1,\overline{x_1}
\end{array}
\right\}
\eeq{piZ}
Any nonempty subset~$Z$ of~$Y$ that is not outbound in~$Y$ for~$\Pi_z$
must contain either~$x_i$ or~$\overline{x_i}$ for every index $1\le i\le n$
of literals in~$Y$.
\begin{lemma} \label{lemma:piZ}
Let $Y = \{x_1,\overline{x_1},\ldots,x_n,\overline{x_n}\}$ be a set of atoms
and $\Pi_z$ as in~(\ref{piZ}).
Any nonempty subset $Z$ of $Y$ that contains neither $x_i$ nor $\overline{x_i}$
for some $1\le i\le n$ is outbound in~$Y$ for~$\Pi_z$.
\end{lemma}
\proof
Let $Z$ be a nonempty subset of~$Y$ that contains 
neither~$x_i$ nor~$\overline{x_i}$ for some index $1\le i\le n$.
For some index $1\le j\le n$, either~$x_j$ or~$\overline{x_j}$ is 
contained in~$Z$. 
Let us fix $i$~and~$j$ such that $j=i-1$ if $i\ge 2$ and $j=n$ if $i=1$.
For one of the rules $x_j \ar x_i,\overline{x_i}$ or
$\overline{x_j} \ar x_i,\overline{x_i}$ of form~(\ref{abf-dis}) in~$\Pi_z$, 
we have
\begin{itemize}
\item $A\cap Z \ne\emptyset$,
\item $A\cap (Y\setminus Z) = \emptyset$,
\item $B\cap Z = \emptyset$, and
\item $B\cap (Y\setminus Z) \ne\emptyset$.
\end{itemize}
That is, $Z$ is outbound in $Y$ for $\Pi_z$.
\qed

The following subprogram constitutes the second part of the reduction:
\beq
\Pi_y =
\left\{\;
\begin{array}{l@{\hspace{15mm}}l}
x_1;\overline{x_1} \ar x_2
& 
x_1;\overline{x_1} \ar \overline{x_2}
\\
x_2;\overline{x_2} \ar x_3
&
x_2;\overline{x_2} \ar \overline{x_3}
\\[0mm]
\;\vdots
&
\;\vdots
\\[0mm]
x_{n-1};\overline{x_{n-1}} \ar x_n
&
x_{n-1};\overline{x_{n-1}} \ar \overline{x_n}
\\
x_n;\overline{x_n} \ar x_1
&
x_n;\overline{x_n} \ar \overline{x_1}
\end{array}
\right\}
\eeq{piY}
Any proper subset~$Z$ of~$Y$ that contains~$x_i$ and~$\overline{x_i}$
for some index $1\le i\le n$ of literals in~$Y$ is outbound in~$Y$ for~$\Pi_y$.
\begin{lemma} \label{lemma:piY}
Let $Y = \{x_1,\overline{x_1},\ldots,x_n,\overline{x_n}\}$ be a set of atoms
and $\Pi_y$ as in~(\ref{piY}).
Any proper subset $Z$ of $Y$ that contains both $x_i$ and $\overline{x_i}$
for some $1\le i\le n$ is outbound in~$Y$ for~$\Pi_y$.
\end{lemma}
\proof
Let $Z$ be a proper subset of~$Y$ that contains 
both~$x_i$ and~$\overline{x_i}$ for some index $1\le i\le n$.
For some index $1\le j\le n$, either~$x_j$ or~$\overline{x_j}$ is not
contained in~$Z$. 
Let us fix $i$~and~$j$ such that $j=i+1$ if $i\le n$ and $j=1$ if $i=n$.
For one of the rules $x_i;\overline{x_i} \ar x_j$ or
$x_i;\overline{x_i} \ar \overline{x_j}$ of form~(\ref{abf-dis}) in~$\Pi_y$, 
we have
\begin{itemize}
\item $A\cap Z \ne\emptyset$,
\item $A\cap (Y\setminus Z) = \emptyset$,
\item $B\cap Z = \emptyset$, and
\item $B\cap (Y\setminus Z) \ne\emptyset$.
\end{itemize}
That is, $Z$ is outbound in $Y$ for $\Pi_y$. 
\qed

Finally, the following subprogram encodes a clause~$c$ in~$F$:
\beq
\Pi_c =
\left\{
\overline{l^1};\overline{l^2} \ar l^3 
\hspace{15mm}
\overline{l^1};\overline{l^3} \ar l^2
\hspace{15mm}
\overline{l^2};\overline{l^3} \ar l^1
\right\}
\eeq{piC}
The subprogram ensures that any noncontradictory set of literals, associated with variables
occurring in clause $c$, is outbound if it contains the complements of
$c$'s literals.
\begin{lemma} \label{lemma:piC}
Let $Y = \{x_1,\overline{x_1},\ldots,x_n,\overline{x_n}\}$
be a set of atoms,
$c = l^1 \vee l^2 \vee l^3$ be a clause,
and $\Pi_c$ as in~(\ref{piC}).
Any proper subset~$Z$ of~$Y$ that 
contains~$\overline{l^1}$,~$\overline{l^2}$, and~$\overline{l^3}$,
but not $l^1$, $l^2$, and $l^3$,
is outbound in~$Y$ for~$\Pi_c$.
\end{lemma}
\proof
Let~$Z$ be a proper subset of~$Y$ that 
contains~$\overline{l^1}$,~$\overline{l^2}$, and~$\overline{l^3}$,
but not~$l^1$,~$l^2$, and~$l^3$.
We consider the case that~$l^1$ is not contained in~$Z$, the cases 
for~$l^2$ and~$l^3$ are symmetric.
For rule $\overline{l^2};\overline{l^3} \ar l^1$ of form~(\ref{abf-dis}) in~$\Pi_c$,
we have
\begin{itemize}
\item $A\cap Z \ne\emptyset$,
\item $A\cap (Y\setminus Z) = \emptyset$,
\item $B\cap Z = \emptyset$, and
\item $B\cap (Y\setminus Z) \ne\emptyset$.
\end{itemize}
That is, $Z$ is outbound in $Y$ for $\Pi_c$. 
\qed

Having described all three parts, we can now complete the reduction of~$F$ to~$\Pi_F$:
\beq
\Pi_F = \Pi_z \cup \Pi_y \cup \mbox{$\bigcup_{c\in F}$}\Pi_c
\eeq{piF}
By Lemma~\ref{lemma:piZ}, any nonempty proper subset~$Z$ of~$Y$ that is not
outbound in~$Y$ for~$\Pi_F$ must contain either~$x_i$ or~$\overline{x_i}$ for
each index $1\le i\le n$ of literals in~$Y$.
Due to subprogram~$\Pi_y$, any such subset~$Z$ of~$Y$ must, by
Lemma~\ref{lemma:piY}, not contain both~$x_i$ and~$\overline{x_i}$ 
for any index $1\le i\le n$ of literals in~$Y$.
Hence, any subset~$Z$ of~$Y$ that is not outbound in~$Y$ for~$\Pi_F$ corresponds
to a total interpretation of variables occurring in~$F$.
Finally, $Z$ must not contain $\overline{l^1},\overline{l^2},\overline{l^3}$
for any clause~$c=l^1 \vee l^2 \vee l^3$ in~$F$ by Lemma~\ref{lemma:piC}.
Hence, some nonempty proper subset~$Z$ of~$Y$ is not outbound in~$Y$ for~$\Pi_F$
iff~$Z$ corresponds to a model of~$F$.
If no such~$Z$ exists, then~$F$ is unsatisfiable, and~$Y$ is an elementary set of~$\Pi_F$.
\qed

}

\bibliographystyle{aaai}

\end{document}